\documentclass[sigconf]{acmart}

\renewcommand\footnotetextcopyrightpermission[1]{} 
\AtBeginDocument{%
  \providecommand\BibTeX{{%
    \normalfont B\kern-0.5em{\scshape i\kern-0.25em b}\kern-0.8em\TeX}}}

\begin{document}

\fancyhead{}

\title{MBRS : Enhancing Robustness of DNN-based Watermarking by Mini-Batch of Real and Simulated JPEG Compression}

\author{Zhaoyang Jia}
\email{jzy_ustc@mail.ustc.edu.cn}
\affiliation{%
  \institution{University of Science and Technology of China}
  \streetaddress{No. 443, Huangshan Road}
  \city{Hefei}
  \state{Anhui}
  \country{China}
  \postcode{230027}
}

\author{Han Fang}
\authornote{Corresponding author.}
\email{fanghan@mail.ustc.edu.cn}
\affiliation{%
  \institution{University of Science and Technology of China}
  \streetaddress{No. 443, Huangshan Road}
  \city{Hefei}
  \state{Anhui}
  \country{China}
  \postcode{230027}
}

\author{Weiming Zhang}
\authornotemark[1]
\email{zhangwm@ustc.edu.cn}
\affiliation{%
  \institution{University of Science and Technology of China}
  \streetaddress{No. 443, Huangshan Road}
  \city{Hefei}
  \state{Anhui}
  \country{China}
  \postcode{230027}
}

\renewcommand{\shortauthors}{Jia, Fang and Zhang}

\begin{abstract}
    Based on the powerful feature extraction ability of deep learning architecture, recently, deep-learning based watermarking algorithms have been widely studied. The basic framework of such algorithm is the auto-encoder like end-to-end architecture with an encoder, a noise layer and a decoder. The key to guarantee robustness is the adversarial training with the differential noise layer. However, we found that none of the existing framework can well ensure the robustness against JPEG compression, which is non-differential but is an essential and important image processing operation. To address such limitations, we proposed a novel end-to-end training architecture, which utilizes \textbf{M}ini-\textbf{B}atch of \textbf{R}eal and \textbf{S}imulated JPEG compression (\textbf{MBRS}) to enhance the JPEG robustness. Precisely, for different mini-batches, we randomly choose one of real JPEG, simulated JPEG and noise-free layer as the noise layer. Besides, we suggest to utilize the Squeeze-and-Excitation blocks\cite{2017Squeeze} which can learn better feature in embedding and extracting stage, and propose a “message processor” to expand the message in a more appreciate way. Meanwhile, to improve the robustness against crop attack, we propose an additive diffusion block into the network. The extensive experimental results have demonstrated the superior performance of the proposed scheme compared with the state-of-the-art algorithms. Under the JPEG compression with quality factor $Q=50$, our models  achieve a bit error rate less than 0.01\% for extracted messages, with PSNR larger than 36 for the encoded images, which shows the well-enhanced robustness against JPEG attack. Besides, under many other distortions such as Gaussian filter, crop, cropout and dropout, the proposed framework also obtains strong robustness. The code implemented by PyTorch \cite{2011torch7} is avaiable in \textcolor[rgb]{0,0,1}{\url{https://github.com/jzyustc/MBRS}}.
\end{abstract}


\maketitle

\begin{figure}[h]
	\includegraphics[width=\linewidth]{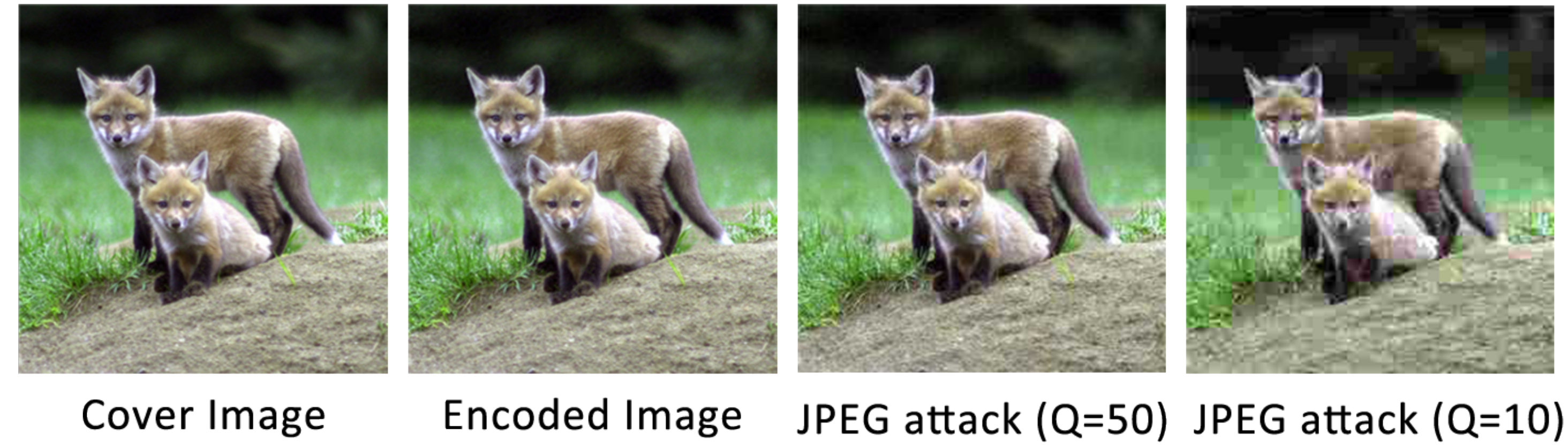}
	\caption{Our model can embed a given secret message into a cover image, to produce an encoded image. The message can be extracted from the compressed encoded images with different quality factors.}
	\Description{A overview of functions of our model.}
\end{figure}

\begin{figure*}[t]
	\includegraphics[width=0.9\linewidth]{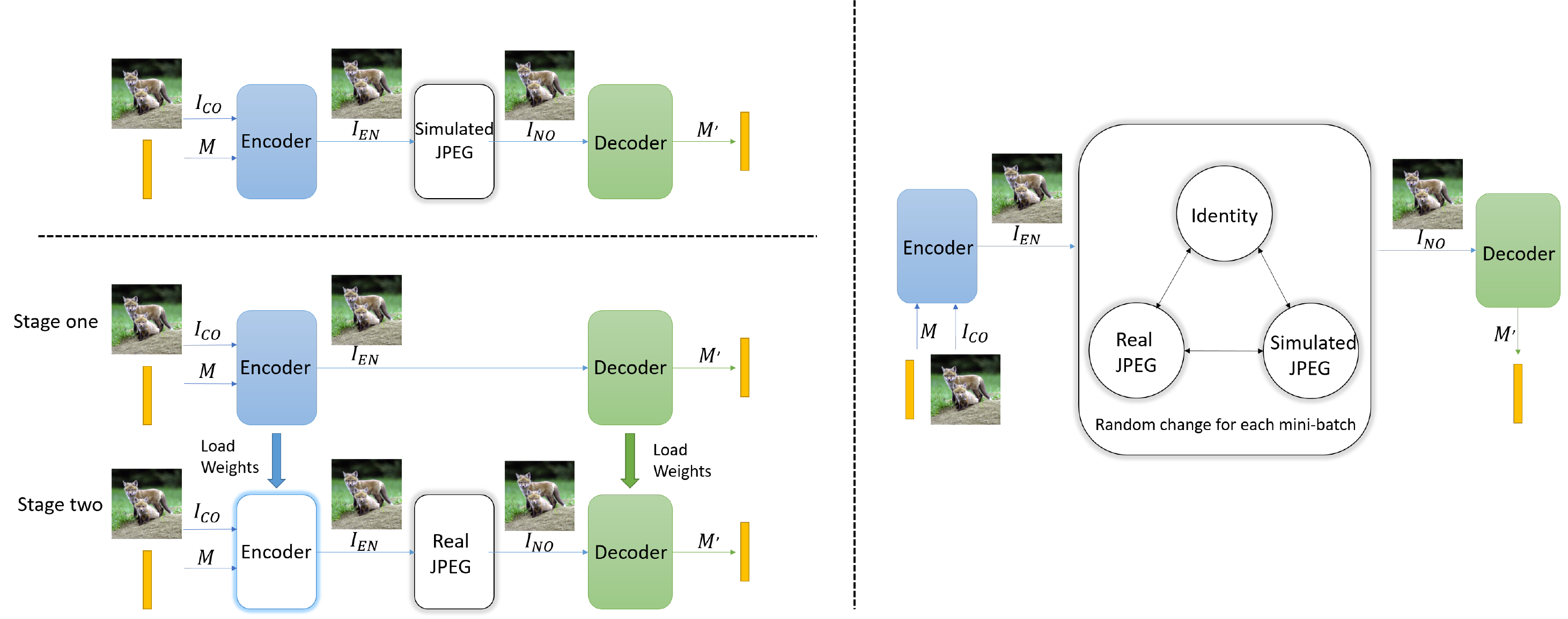}
	\caption{The encoder-decoder structure is widely used in DNN-based watermarking models. The left top image shows the One-stage End-to-end training with Differential Simulated JPEG compression (OEDS) method, where the encoder and decoder are jointly trained with simulated noise layer. The left bottom image shows the Two-stage Separable framework with Real JPEG compression (TSR) proposed in \cite{inproceedings}, where the encoder and decoder train together without noise in stage one, and train the decoder alone with real JPEG noise in stage two. The right image is our Mini-Batch of Real and Simulated JPEG compression (MBRS) method, which random changes the noise layer from the simulated JPEG, real JPEG and noise-free (called Identity) layer for each mini-batch.}
	\Description{A Comparison among OEDS, TSR and MBRS method.}
\end{figure*}

\section{Introduction}

Digital watermarking was first introduced and defined by Ron van Schyndel in 1994 \cite{413536}, and then was widely used for the protection of intellectual property of images \cite{817598,736686}, videos \cite{650120,879337} and audios \cite{SWANSON1998337,7089781}. Specifically, the digital image watermarking aims to embed the secret message in an invisible method, and then extract them from the encoded image even if the image is modified, which means that both image quality and robustness are required. The earliest research \cite{413536} encodes the secret message on the least significant bits (LSB) of the image pixels, but this method can be easily detected by statistical measures \cite{2002Practical,ISI:000183618000029}. Then, the researchers turned to focus on the frequency domain, and they found it much more robust to encode messages in DCT domain \cite{2018Hybrid} and DWT domain \cite{2002Digital}. However, all those traditional methods rely heavily on shallow hand-craft image features, which indicates that they are not able to make full use of the cover images, therefore have many limitations in robustness. In recent years, with the development of deep learning, many DNN-based models \cite{Zhu_2018_ECCV,2018ReDMark,inproceedings,2019StegaStamp} have been applied in digital image watermarking for stronger robustness. Zhu et. al \cite{Zhu_2018_ECCV} proposed a DNN-based auto-encoder to jointly train the encoder and decoder with a noise layer. It succeed in both image quality and robustness of various distortions, and over-performs most of traditional methods greatly.

However, most of the DNN-based watermarking frameworks don't perform well in the robustness against JPEG compression\cite{125072}, which is the most common lossy-compression method on the Internet. In JPEG compression, we need to quantize the DCT coefficients of image blocks, and this step is non-differential, which makes the gradients propagated passed \cite{ISI:A1986E327500055} to the encoder become zero. In this case, the encoder cannot update the parameters according to the decoding loss, leading to the wrong direction of enhancing robustness. Many methods have been proposed to solve this non-differential problem. Some of them \cite{Zhu_2018_ECCV,2018ReDMark,2019StegaStamp} utilize the \textbf{O}ne-stage \textbf{E}nd-to-end training with \textbf{D}ifferential \textbf{S}imulated JPEG compression (\textbf{OEDS}) method to meet the request of back-propagation gradients. But the results are not satisfactory because of the natural lack of replacing real noise with simulations. A \textbf{T}wo-stage \textbf{S}eparable framework with \textbf{R}eal JPEG compression (\textbf{TSR}) \cite{inproceedings} is proposed to separate the training process of the encoder and the decoder, so as to eliminate the differential limitation of noise layer. In this method, the model is jointly trained without noise in stage one, and then the decoder is trained alone under real JPEG compression. This is a fantastic idea to solve the problem, but the encoder also can’t get information about decoding with JPEG distortion, as a result the performance is not good enough. 

Both OEDS and TSR don’t perform well under JPEG compression, but we can enhance the robustness by combining the advantages of 1) OEDS to jointly train the model, and of 2) TSR to train the decoder with real JPEG compression. However, it’s a difficult task to merge those two methods in an appropriate way. An intuitive approaches is to replace noise-free layer with simulated JPEG in stage one of TSR and maintain the stage two. Unfortunately, this method still has a poor performance against JPEG compression, which can be interpreted as the two-stage strategy can only find the optimal solution for each single stage instead of the global optimal solution.

To address such limitations of above methods, in this paper, we propose our \textbf{M}ini-\textbf{B}atch of \textbf{R}eal and \textbf{S}imulated JPEG compression (\textbf{MBRS}) method. As is shown  in Figure 2, for each mini-batch, we change the noise layer randomly from real JPEG, simulated JPEG and noise-free layer, so that different mini-batches can train the model for different purpose: 1) real JPEG help the decoder to obtain robustness under the JPEG compression, 2) simulated JPEG train the encoder-decoder jointly, and 3) the noise-free layer makes sure the decoding ability without compression. The frequent switching of noise layer in each mini-batch helps the model to search the solution in different directions, which guarantees the optimal solution for the whole task. Although in some batches the non-differential real JPEG is applied for end-to-end training, we can use momentum-based updating optimization method to ensure the correctness of the whole updating direction.

What’s more, we present an auto-encoder based network to make the best use of the MBRS method. JPEG compression can be demonstrated as a kind of limitation of high frequency part of DCT coefficients, so we utilize Squeeze-and-Excitation(\textbf{SE}) blocks \cite{2017Squeeze} to learn features from the frequency domain. And we also propose a block denoted as a “message processor” to auto-learn a better way to expand the secret messages and achieve redundancy. In this process, we use the trick of strength factor to adjust the trade-off between image quality and robustness. Experiments shows that our architecture achieves a high image quality, and the decoding bit error rate is reduced to almost 0\%. And we can also use a similar way to train a model that is robust against many other distortions, such as Gaussian filter, Dropout, etc.

Lastly, in order to enhance robustness against crop and cropout attack, we add a diffusion block and an inverse diffusion block to diffuse the secret message into the whole image. As a result, the model can be much more robust against crop and cropout attack, while the decoding accuracy under other noises won't decrease too much. 

In summary, we list the contribution in this paper as below :

\begin{itemize}
	\item [1.] 
	We introduce a novel MBRS training method to enhance the robustness against JPEG compression greatly.    
	\item [2.]
	We propose an auto-encoder based network utilizing SE blocks and a message processor for higher embedding and extracting ability.
	\item [3.]
	As a supplement, we propose an independent diffusion block and inverse diffusion block for robustness against crop attack.
\end{itemize}

The reminder of the paper is arranged as following. In Section 2 we review some related works about the DNN-based watermarking frameworks and methods to approximate JPEG. The details of the proposed architecture are introduced in Section 3. The results from experiment are presented in Section 4. And finally we make a conclusion in Section 5.

\section{RELATED WORKS}

\emph{Deep Learning for Digital Watermarking}. Recently, many deep learning based watermarking framework are proposed, which greatly utilize the powerful feature extraction ability of neural network architecture. For example,  Zhu et. al. \cite{Zhu_2018_ECCV} proposed an end-to-end DNN-based model for watermarking. The main architecture is an auto-encoder like encoder-noise layer-decoder structure, and a discriminator is used for more realistic visual effect. Ahmadi et. al \cite{2018ReDMark} proposed a framework which supports operating on different domains like DCT domain, and it uses a residual structure for the encoder with a strength factor to control the strength of watermark patterns in the image. And Tancik et. al \cite{2019StegaStamp} focuses on a specific robustness: print-shooting robustness. To achieve such special robustness, they simulated the print-shooting process with several differential operation and apply them in the noise-layer. 

Although the end-to-end framework facilitated the joint learning of encoder and decoder, the differential limitation of the noise layer makes it inapplicable in practice. As a result, a two-stage separable deep learning framework \cite{inproceedings} is proposed for practical watermarking, the encoder and decoder are initialized without noise layer in stage one and the decoder will be enhanced alone by non-differential distortions in the stage two. 

All these deep learning based methods achieve great performance in terms of image quality and some image processing robustness, but none of them can well deal with JPEG compression distortion. Since JPEG compression is a non-differential process, the end-to-end framework cannot be directly applied, and under OEDS method the simulated distortion cannot guide the correct updating direction of model. Besides, for the two-stage framework with TSR method, the JPEG compression may greatly influence the encoder feature, so that even with adversarial training, the decoder still cannot extract enough feature for decoding. In fact, the two-stage method is similar to a Greedy algorithm, so it may find the local optimal results but cannot find the global optimal results.

\begin{figure*}[t]
	\includegraphics[width=\textwidth]{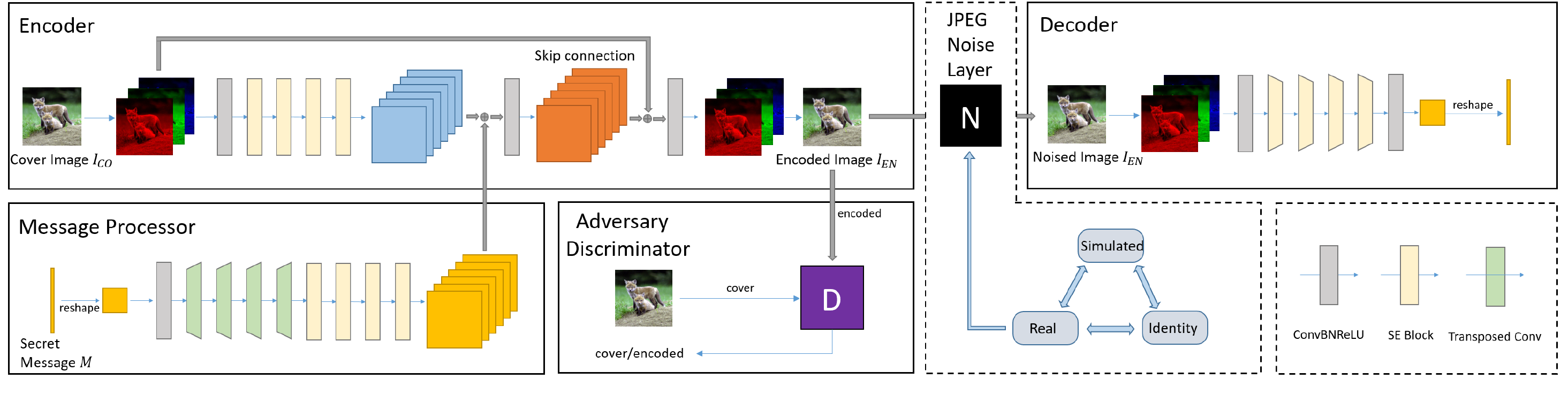}
	\caption{Model overview. The message processor auto-learns methods for expanding message and realizing redundancy; the encoder with SE blocks embeds the secret message into the cover image; the JPEG noise layer change the kind of noise according to MBRS method; and the decoder extracts the secret message from the embedded image. An additional adversary discriminator is used to distinguish the cover image and the embedded image. }
	\label{fig:teaser}
\end{figure*}

\emph{Adversary Networks}. The adversarial training was first introduced by Goodfellow in 2014 \cite{ISI:000452647101094} to estimate generative models. And many improvements are proposed to generate lots of variants of GAN, such as DCGAN \cite{2015Unsupervised} and WGAN \cite{2017Wasserstein} to improve the stability of training and quality of generated images, CycleGAN \cite{2017Unpaired} and pix2pix \cite{2016Image} models for image to image translation, and CGAN \cite{2014Conditional} to add more conditions for image generating. Many models for watermarking also use an adversary for the encoder to obtain higher image quality, and they all get excellent results. \\\par

\emph{JPEG simulations}. For the encoder-decoder structure in image watermarking, to meet the requirement of the one-stage end-to-end training, many methods have been proposed to use several differential operations to simulate the JPEG compression. In \cite{Zhu_2018_ECCV}, Zhu et. al proposed a method named \textbf{JPEG-Mask}, which zeros a fixed set of high frequency coefficients, and only maintain the $5\times5$ low frequency region of Y channel, $3\times3$ low frequency region for U and V channels. By utilizing this method, the network can obtain JPEG robustness in some extend. And some other researches focus on the simulation of the quantization function in JPEG compression. Shin and Song \cite{2017jpeg} only approximate the quantization step near zero, with the piece-wise function 
\begin{equation}
    q(x)=
    \left\{
    \begin{aligned}
    & x^3 &: |x| < 0.5 &\\
    & x   &: |x| \ge 0.5 &
    \end{aligned}
    \right.
\end{equation}
We denoted the method in \cite{2017jpeg} as \textbf{JPEG-SS} in the following sections.  \\\par

\section{PROPOSED FRAMEWORK}

\subsection{Model Architecture}

Our goal is to train an end-to-end watermarking model that is robust to JPEG compression. As showed in Figure 3, this model architecture includes five components : 1) 	Message Processor MP with parameters $\theta_M$ receives the binary secret message $M\in\{0,1\}^L$ of length $L$, and outputs a message feature map  $M_{en}\in \textbf{R}^{C'\times H\times W} $, where $C'$ is the channel number of the feature map.2) Encoder $E$ with parameters $\theta_E$ receives the RGB cover image $I_{co}$ in the shape of $3\times H\times W$ and the message feature map $M_{en}$ as input, and product the encoded image $I_{en}$ of shape $3\times H\times W$. 3) Noise Layer $N$ randomly selects noise according to the MBRS method. It receives $I_{en}$ and outputs the noised image $I_{no}$ of the same shape. 4) Decoder $D$ with parameters $\theta_D$ recovers the secret message $M'$ of length $L$ from the noised image $I_{no}$ . 5) Adversary discriminator $A$ with parameters $\theta_A$ receive the image  $I_{en}$ or $I_{co}$ to predict the probability of a given image being encoded. \\

\textbf{\emph{Message Processor}}. In order to better support the encoding process, the message should be redundant and expanded in a more appropriate approach. To this end, we add the message processor to process the message and then feed the feature map to the encoder. In the first step, the secret message $M$ with a length of $L$ is reshaped to ${0,1}^{1\times h\times w}$ where $L=h\times w$. It is amplified by a single $3\times3$ convolution layer, followed by batch normalization and ReLU activation (denoted as ConvBNReLU), and then expanded to $C\times H\times W$ ($C$ is the feature channel number, and $H,W$ are the height and width of the cover images) by several transposed convolution layers with stride = 2. After that, the features of the "message image" are extracted by several SE blocks that maintains the shape. The message feature map is then sent to the encoder to be concentrated with the image features.

During the expanding step, because each transposed convolutional layer causes the width and height of the input tensor to double, the secret message's length $L$ and the cover image's shape $H\times W$ are generally required to follow this relationship : 
$$
L=h\times w=(H/2^n)\times(W/2^n)
$$
where $n\in Z^*$ is an integer decided by $L$, $H$ and $W$.\\

\textbf{\emph{Encoder}}. The encoder aims to encode the watermark into the host image with low visual distortion. So such task should be done with a fully understanding of the image. To learn better image features, we utilize the structure of SE block which is widely used in feature learning tasks as the basic block. Firstly, we amplify the cover image $I_{CO}$ of shape $3\times H\times W$ through a $3\times3$ ConvBNReLU layer then extracts image features of the same shape with several SE blocks. Obtaining the features of the cover image and the message feature map from the message processor, the encoder simply concentrates them and then mapped through a $3\times3$ ConvBNReLU layer. Finally, we concentrate the obtained tensor and the cover image to a new tensor , and it's fed into a $1\times1$ convolutional layer to obtain the encoded image $I_{EN}$.

The object of encoder training is to minimize the $L_2$ distance between $I_{CO}$ and $I_{en}$ by updating parameters $\theta_E$ to make them the visually similar: 
$$
L_{E_1}=MSE(I_{co},I_{en})=MSE(I_{co},E(\theta_E,I_{co},M))
$$

\textbf{\emph{Decoder}}. In the decoder, the noised image $I_{no}$ is also amplified by a $3\times3$ ConvBNReLU layer, and then turned into $C\times h\times w$  by several SE blocks. Finally, we use a $3\times3$ convolutional layer to convert the multi-channels tensor into 1-channel, and reshape it to get the decoded message $M'$. 

The object of decoder training is to minimize the $L_2$ distance between $M$ and $M'$ by updating parameters $\theta_D$ to make them the same : 
$$
L_D=MSE(M,M')=MSE(M,D(\theta_D,I_{no}))
$$ \\

\textbf{\emph{Adversary}}. The adversary discriminator is simply consists of several $3\times3$ convolutional layers and a global average pooling layer. The discriminator performs as an adversary of the encoder, which means it needs to distinguish the encoded image while the encoder should prevent it from doing this: 

Update parameters $\theta_A$ to minimize
$$
L_A=\log(1-A(\theta_A,E(\theta_E,I_{co},M)))+\log(A(\theta_A,I_{co}))
$$

And update parameters $\theta_E$ to minimize 
$$
I_{E_2} = \log(A(\theta_A,I_{en}))=\log(A(\theta_A,E(\theta_E,I_{co},M)))
$$ \\
In total, the target loss function is $L=\lambda_E L_{E_1}+\lambda_D L_D+\lambda_A L_{E_2}$ for the encoder and decoder, and loss $L_A$ for the adversary discriminator.

\subsection{Noise Layer}

\textbf{\emph{Real JPEG Compression in Encoder-Decoder Structure}}. As mentioned in Section 1, in JPEG compression, we need to quantize the DCT coefficients according to the quantization tables, but this process is non-differential. That is, the gradient propagated back will be zero, so the decoding loss that used to guide the encoder will also be zero: $\frac{\partial  L_D}{\partial \theta_E}=0$. And this means the encoder cannot update the parameters according to the decoding results. So that the whole architecture will behaved as: Encoder only tries to generate better image, and decoder tries to recover the message, but not jointly trained for better performance, which may resulted in the best visual quality but worst robustness since the encoder may not realize the embedding process at all.\\

\textbf{\emph{Our MBRS Methods}}. To overcome such limitations, we use a novel Mini-Batch of Real and Simulated JPEG compression (MBRS) method in our model. In each mini-batch, we random choose one from a real JPEG compression layer, a differential simulation of JPEG layer and a noise-free layer (called Identity) as the noise layer in training. We believe such design will help the model to find the global optimal solution. The functionality of each operation MBRS method are demonstrated as:
\begin{itemize}
    \item [1.] 
    Real JPEG Layer: Real JPEG Layer is used for decoder to learn enough feature for decoding even after real JPEG Compression.  
    \item [2.] 
    JPEG Mask Layer: As a simulation of JPEG compression, JPEG-Mask provide the gradient that can be propagated back to the encoder, so that not only the encoder and the decoder can be both joint-trained. 
    \item [3.] 
    Identity: The existence of identity layer ensures the decoding ability without JPEG compression.
\end{itemize}
And by updating in different directions for each mini-batch, we except to obtain the global optimal solution.\\

\textbf{\emph{Why the combined noise-layer works}}. As mentioned in the previous paragraph, only using real JPEG for iterations will lead to a lack of robustness, but with the MBRS method, such limitations can be greatly reduced. The reason can be explained as: the mini-batch with real JPEG is usually applied after a mini-batch with JPEG Mask or Identity. So even the encoder is not updated according to the decoding loss, the encoder is still updated with the momentum that with the former mini-batch, which ensures the correctness of the direction. So in such mini-batch, even the encoder and decoder is not joint trained, they are still updating in the right way. That is, encoder tries to create better image not only with high visual quality, but also with robustness features, and decoder tries to recover the feature after real JPEG compression. And with the real JPEG layer, the "overfitting" of the simulated JPEG attack can be prevented too. To support this process, we can use momentum-based updating methods like Adam in optimization.\\

\textbf{\emph{Different from TSR}}. The difference between MBRS and TSR can be summarized as: In TSR, there is only one switch from stage one with identity noise layer to stage two with real JPEG. But in our model, the "stage" changes randomly for each mini-batch. In other words, TSR uses a Greedy algorithm to find partial optimal solutions for each stage. As is known, a Greedy algorithm cannot get the best result in most situations. Our MBRS method is designed to search the global optimal solution in the entire solution space step by step in different directions for each mini-batch, so it is more likely to get a better solution. 

\subsection{Strength Factor}

We can get the residual signal of the encoded image and cover image $I_{Diff}=I_{EN}-I_{CO}$, then adjust the trade-off between the quality of encoded images and the recovery messages' accuracy by a strength factor $S$ : $I_{EN,S}=I_{CO}+S\cdot I_{Diff}$. Note that this trick is used only in the encoder, and in decoding process we will not use the cover image. That is, we apply a blind watermarking method in the architecture.

\begin{figure}[h]
	\includegraphics[width=\linewidth]{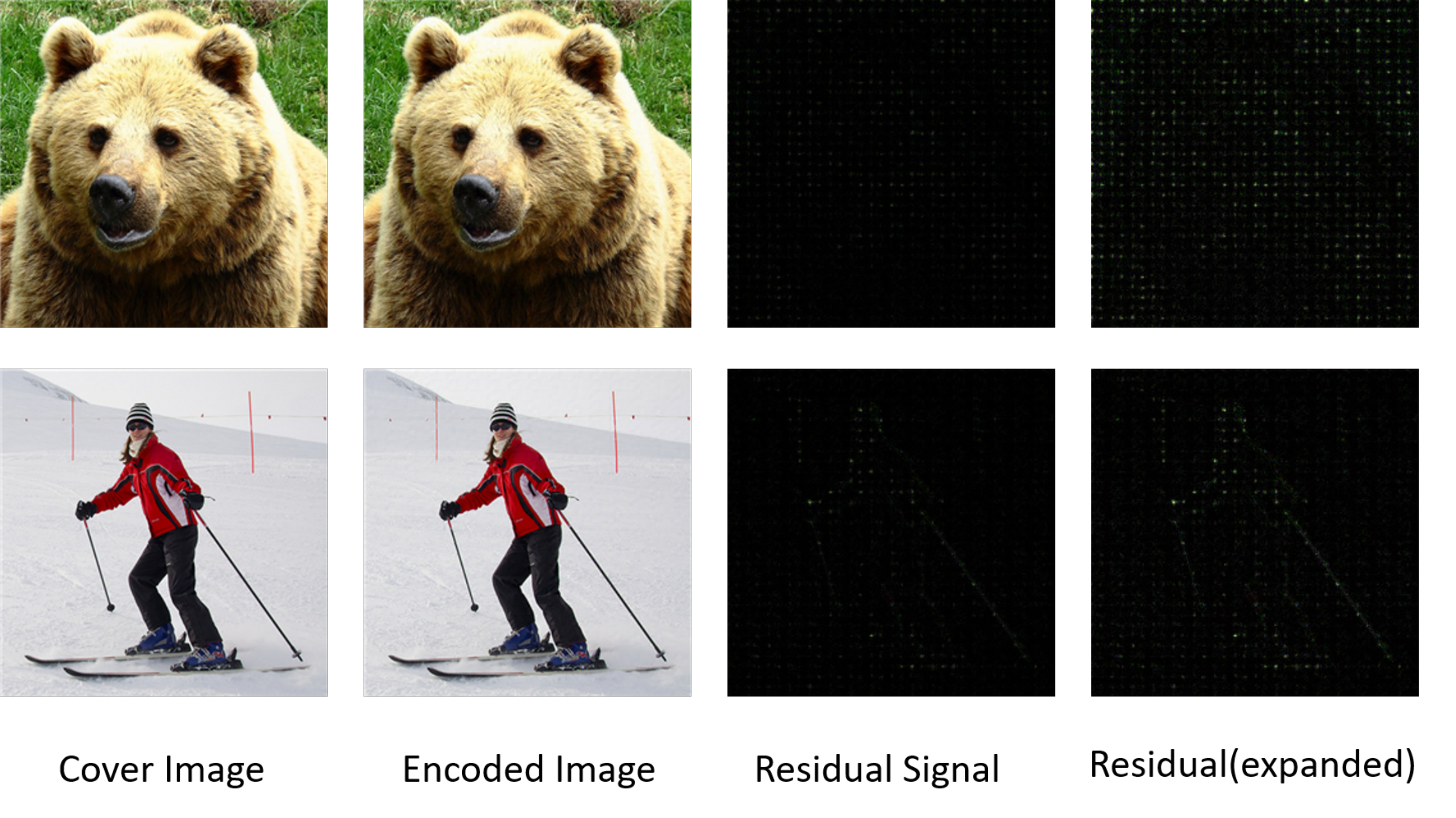}
	\caption{Choose two images from the validation dataset randomly to show the image quality. We test them with JPEG ($Q$=50), and result shows that (top) BER = 0.0\%, PSNR = 40.13 and SSIM = 0.9802, (bottom) BER = 0.0\%, PSNR = 41.26 and SSIM = 0.9491. From left to right are: the cover image $I_{CO}$, the encoded image $I_{EN}$, the residual signal of them $R=|I_{EN}-I_{CO}|$, the normalization of residual signal $R_M=(R-min(R))/(max(R)-min(R))\times255$ . }
	\Description{Results showing.}
\end{figure}

\begin{table*}[t]
	\centering
	\caption{Comparison with the SOTA results. We train our models with MBRS of JPEG-Mask, real JPEG ($Q$=50) and Identity. We test \cite{2019StegaStamp} with the open source pre-trained model, while directly use results reported in \cite{Zhu_2018_ECCV} and \cite{inproceedings} for comparison. However, SSIM is not reported in \cite{Zhu_2018_ECCV} and \cite{inproceedings}, for which we empty these items. PSNR is measured for RGB channels, except in \cite{Zhu_2018_ECCV} they use Y channel of YUV color space.}
	\begin{tabular}{c|c|c|c|c|c}  
		\hline
		\hline
		Model          & \cite{Zhu_2018_ECCV} (specified) & \cite{inproceedings} (specified) & Ours& \cite{2019StegaStamp} & Ours \\
		\hline
		Image size     & $128\times128$ & $128\times128$  & $128\times128$ & $400\times400$  & $400\times400$                     \\
		message length & 30                & 30              & \textbf{64}       & 100        & \textbf{625}                     \\
		Noise Layer    & JPEG-Mask         & JPEG            & Mixed(MBRS)                     & JPEG-SS    & Mixed(MBRS)                      \\
		PSNR           & (for Y) 30.09     & 33.51           & \textbf{36.49}            & 32.36      & \textbf{39.32}                   \\
		SSIM           & -                 & -               & \textbf{0.9173}           & 0.8506     & \textbf{0.9504}                  \\
		BER            & 15\%               & 22.3\%           & \textbf{0.0092\%}         & 0.2418\%     & \textbf{0.0012\%}                 \\
		\hline
		\hline
	\end{tabular}
\end{table*}

\begin{table*}[t]
	\centering
	\caption{Results of robustness against other distortions. Our model with the original structure is not robust enough against crop, cropout and dropout, so we add an additive diffusion block, and then make a comparison with model from \cite{Zhu_2018_ECCV} an \cite{inproceedings} trained by combined noise layer. Strength factor is adjusted for comparison under $PSNR=33.5$.}
	\begin{tabular}{c|c|c|c|c|c|c|c}  
		\hline
		\hline
		Noise & message length & Identity &  Cropout ($p=0.3$) & Dropout ($p=0.3$) &  Crop ($p=0.035$) & GF ($\sigma=2$) & JPEG ($Q$=50) \\
		\hline
		Ours(w/o diffusion block) & 64 & 0\%  & 32.86\%   & 8.12\%  & 45.86\% & 0.0032\% & 4.14\%   \\
		Ours(w/ diffusion block) & 30 & 0\%  & 0.0027\% & 0.0087\% & 4.15\% & 0.011\% & 4.48\%  \\
		\cite{Zhu_2018_ECCV} (combined) & 30  & 0\% & 6\%  & 7\%  & 12\% & 4\% & 37\% \\
		\cite{inproceedings} (combined) & 30  & 0\% & 2.7\%  & 2.6\%  & 11\% & 1.4\% & 23.8\% \\
		\hline
		\hline
	\end{tabular}
\end{table*}

\section{EXPERIMENTS}

\textbf{\emph{Implementation Details}}. The model is trained on 10,000 images from the ImageNet dataset, and evaluated on a 5000 images test set from COCO dataset to ensure the generalization of the trained model. The whole framework is implemented by PyTorch\cite{2011torch7} and executed on NVIDIA RTX 2080ti. Messages are sampled randomly at each bit. We train the model with the proposed MBRS method, and test with the JPEG compression function in PIL package in Python. Strength Factor is set as 1 during training. For the weight factors of the loss function, we choose $\lambda_E=1, \lambda_D=10$ and $\lambda_A=0.0001$. And for gradient descent method, we use Adam with a learning rate of $10^{-3}$ and default hyperparameters. The size of mini-batch is $16$, and the model is trained for $100$ epochs.\\

\textbf{\emph{Metric}}. There are two main indicators to judge our model : \emph{robustness}, measured by Bit Error Rate (BER) of decoded message; and \emph{image quality}, measured by PSNR \cite{2010Stego} and SSIM \cite{2004Image}.\\

\textbf{\emph{Baseline}}. Our baseline for comparison are \cite{Zhu_2018_ECCV}, \cite{inproceedings} and \cite{2019StegaStamp}. All of these works are deep learning based watermarking schemes. The authors of \cite{2019StegaStamp} open source of both their codes and their model, so we use the pre-trained model for comparison.  We also try to conduct experiments of \cite{Zhu_2018_ECCV} and \cite{inproceedings}, but cannot achieve the best performance as they have reported.  So in order to respect the results they have reported, we directly use the results published in \cite{Zhu_2018_ECCV} and \cite{inproceedings}.

\subsection{Visual Quality}

We train the basic model of our architecture and show the visual quality of encoded images. The basic model is trained with JPEG-Mask, real JPEG ($Q=50$), and Identity layer with mini-batch strategy. The visual quality of the encoded images is shown in Figure 4, the residual signal $R=|I_{EN}-I_{CO}|$ and the normalized residual image $R_M=(R-min(R))/(max(R)-min(R))\times255$ are also shown. 
We can see that the proposed encoder can adaptively embeds the message into the cover image while maintains the high visual quality. Besides, from the residual signal, we can see that the texture complex region of the image is embedded with more message information which greatly improve the transparency.

\subsection{Comparison with Previous Methods}

In this section, we compare our model with three SOTA models, \cite{Zhu_2018_ECCV}, \cite{inproceedings} and \cite{2019StegaStamp}. Since the input image size message length of each methods varies, for a fair comparison, we train our model with the same image size and the similar message length. Specifically, we use $L=64$ for $3\times128\times128$ images, and $L=625$ for $3\times400\times400$ images, both of which have higher message capacity than the original articles. The detailed information is shown in Table 1.

\subsubsection{JPEG Compression}

In this section, we mainly show and discuss the JPEG robustness of the proposed method. The test model is trained by MBRS method with JPEG-Mask, real JPEG with ($Q=50$), and Identity layer. All the testing process is carried out under real JPEG compression with $Q=50$. For \cite{Zhu_2018_ECCV} and \cite{inproceedings} we use the results of the model that is specified for JPEG compression. As shown in Table 1, our model can not only maintain higher image quality but also lower BER. In particular, our model achieves the BER that is less than $0.01\%$, which shows that our architecture enhance the robustness against JPEG compression evidently.

\subsubsection{Other Distortion}

In addition to JPEG compression distortion, the proposed arcchitecture can be also used for other image processing distortions such as Gaussian Filter, crop, cropout, dropout, etc. We train a model to embed 64 bits message into $128\times128$ image with mini-batch strategy and the noise layer is consist of JPEG-Mask, real JPEG ($Q$=10), Identity, Gaussian Filter (GF, $\sigma=2$) and Crop ($p=0.0225$). Since our decoder is designed for extracting from a fixed size of images, we pad the cropped images with pixels expressed as (255, 255, 255) in RGB color space. We test the model with Identity, Cropout ($p=0.3$), Dropout ($p=0.3$), Crop ($p=0.035$), Gaussian Filter ($\sigma=2$) and real JPEG compression ($Q$=50), each time for one kind of distortions. As is shown in Table 2, to compare with model trained from \cite{Zhu_2018_ECCV} and \cite{inproceedings}, we find that our model over-performs their model under Identity, JPEG compression and Gaussian filter distortions. However, our model also shows the weakness towards the cropout and crop attack, which accounts for the special structure of our network. To solve this problem, we add an additive diffusion and inverse-diffusion block into the model, and result in Table 2 shows that the model with diffusion block successfully obtains robustness again crop, dropout and cropout. More detailed demonstrations and the structure of diffusion block will be discussed in Section 4.3.5.

\begin{table*}[t]
	\centering  
	\caption{BER, PSNR and SSIM value under several strength factors. BER is tested under different quality factor of JPEG compression. For different applications, we can adjust strength factor to obtain a BER of 0.00\% with PSNR=34, or a much higher image quality with PSNR=42 and BR=1.35\%.}  
	\begin{tabular}{c c|c|c|c|c|c|c|c}  
		\hline
		\hline
		\multicolumn{2}{c|}{Strength factor} & 0.6 & 0.8 & \textbf{1.0} & 1.2 & 1.4 & 2.0 & 2.4   \\
		\hline
		& Q=10 & 40.74\%  & 36.91\%  & \textbf{33.17\%}  & 29.50\%  & 26.19\%  & 18.26\%  & 14.49\% \\
		& Q=30 & 19.76\%  & 12.64\%  & \textbf{8.58\%}  & 5.90\%  & 4.01\%  & 0.13\%  & 0.015\% \\
		BER & Q=50 & \textbf{11.81\%}  & \textbf{5.27\%}  & \textcolor{blue}{\textbf{1.35\%}}  & \textbf{0.17\%}  & \textbf{0.036\%}  & \textbf{0.00024\%}  & \textcolor{red}{\textbf{0.00\%}} \\
		& Q=70 & 2.37\%  & 0.14\%  & \textbf{0.0098\%}  & 0.0017\%  & 0.00024\%  & 0.00\%  & 0.00\% \\
		& Q=90 & 0.51\%  & 0.020\%  & \textbf{0.00063\%}  & 0.00016\%  & 0.00\%  & 0.00\%  & 0.00\% \\
		\hline
		\multicolumn{2}{c|}{PSNR} & 46.48 & 43.98 & \textcolor{blue}{\textbf{42.04}} & 40.46 & 39.12 & 36.02 & \textcolor{red}{\textbf{34.44}}\\
		\hline
		\multicolumn{2}{c|}{SSIM} & 0.9890 & 0.9808 & \textbf{0.9707} & 0.9589 & 0.9459 & 0.9012 & 0.8690 \\
		\hline
		\hline
	\end{tabular}
\end{table*}

\begin{table*}[t]
	\centering
	\caption{Test with different quality of JPEG varying from 10 to 90, and
	two kinds of simulated JPEG compression. PSNR is adjusted to nearby 39 by strength factor for fair. Decoding BER is shown in the table.}
	\begin{tabular}{c c|c c|c c c c c c}  
		\hline
		\hline
		\multicolumn{2}{c|}{Simulated JPEG Compression}& \multicolumn{2}{|c|}{Q of training JPEG}& \multicolumn{6}{|c}{Q of testing JPEG} \\
		\hline
		JPEG-Mask & JPEG-SS &  Q=50 & Q=10 & w/o JPEG & Q=90 & Q=70 & Q=50 & Q=30 & Q=10 \\
		\hline
		$\surd$ &         & $\surd$ &         & \textbf{0.00\%} & \textbf{0.00\%} & \textbf{0.00\%} & \textbf{0.027\%} & 4.13\% & 26.32\% \\
		$\surd$ &         &         & $\surd$ & 0.00040\% & 0.0076\% & 0.085\% & 0.19\% & \textbf{1.21\%} & \textbf{14.55\%} \\
		\hline
	    & $\surd$ & $\surd$ &         & \textbf{0.0\%} & \textbf{0.00024\%} & \textbf{0.0029\%} & \textbf{0.012\%} & \textbf{0.33\%} & 15.79\% \\
        & $\surd$ &         & $\surd$ & 0.0\% & 0.27\% & 0.42\% & 0.63\% & 1.11\% & \textbf{12.32\%} \\
		\hline
		\hline
	\end{tabular}
\end{table*}

\subsection{Ablation Study}

In this section, we mainly conduct the ablation experiments to better illustrated the proposed architecture.

\subsubsection{Strength Factor}

Strength factor is the parameter used for balance the robustness and transparency. To choose the best strength factor for experiment, we set the value of strength factor $S$ from 0.1 to 3.0, with an interval of 0.1, and test the model under different quality factor for JPEG compression. The results are shown in Table 3. As can be seen, with the increment of $S$, PSNR and SSIM values decrease, which indicates the visual quality becomes worse while the extraction accuracy becomes higher. For different performance requirements of application, we can choose different $S$, and in this paper, we adjust the value of $S$ to obtain similar visual quality of different models for fair comparison.

\subsubsection{Quality Factor and Simulated JPEG in Training}

For all JPEG compression above, we choose quality factor $Q = 50$ for real JPEG in training. We also try to train with a much lower $Q = 10$ for comparison, to show how the quality of compression during training influences the performance. And we also train with another kind of simulated JPEG compression: JPEG-SS in training for comparison. The results are shown in Table 4. JPEG with $Q=10$ is a much stronger noise than with $Q=50$, but our model still obtains robustness toward it. In fact, models trained under $Q=10$ perform better than under $Q=50$ against low compression quality but perform worse against high quality, which shows the trade-off between robustness against strong and weak distortions. And model trained with JPEG-Mask also shows better performance than JPEG-SS for high compression quality while performs worse for low quality factor. Think of the fact that we hardly use JPEG compression with Q less than 50 (which affect the visual quality too much), we apply JPEG-Mask and real JPEG with $Q=50$ in training for better performance against high quality compression.

\begin{table}[t]
	\centering
	\caption{The Ablation experiment of selected noise layer. For most cases, we use the Strength Factor to adjust to $PSNR\approx37.8$ for fair. BER is tested under $Q=50$. Results shows in both the combined noise layer JPEG-Mask+JPEG ($Q$=50)+Identity has the best performance.}
	\begin{tabular}{c|c|c|c}  
		\hline
		\hline
		JPEG-Mask & JPEG ($Q$=50) &  Identity  & BER  \\
		\hline
		$\surd$ &         &          & 35.65\%  \\
		$\surd$ &         & $\surd$  & 36.18\%  \\
		        & $\surd$ & $\surd$  & 0.0084\% \\
		$\surd$ & $\surd$ &          & 0.0991\% \\
		$\surd$ & $\surd$ & $\surd$  & \textbf{0.0040\%}  \\
		\hline
		\hline
	\end{tabular}
\end{table}

\subsubsection{Real, Simulated, and Identity Combination}

In our MBRS method, the real JPEG compression, Simulated JPEG-Mask compression and Identity layer are used for different purpose. To better indicate the necessity of each part, we conduct an ablation experiment, in which we remove one of the noise at each training and test for comparison with $Q=50$. As showed in Table 5, absence of any part in MBRS results in the decreasing of robustness, and the existence of real JPEG compression improves the performance greatly. Moreover, we can demonstrate the function of real JPEG layer to prevent the "overfitting" of JPEG-Mask, as mentioned in Section3.2. If we only use JPEG-Mask, the model will overfit the simulation noise, which means it can reach high decoding accuracy for JPEG-Mask, but cannot work well for real JPEG. And since the decoding loss has been too low to be reduced, the model cannot enhance robustness against real JPEG during the later training. But if we add real JPEG to the MBRS method, the decoding loss can be low enough only when the decoder is robust enough against the real JPEG, so the decoding accuracy will not be limited by "overfitting" of simulated JPEG.

\begin{table}[t]
	\centering
	\caption{Comparison among OEDS, TSR, TSR-S and our MBRS method. PSNR is adjusted to about 40 by strength fator for fair. Result shows the advantage of our MBRS method to greatly enhance the robustness.}
	\begin{tabular}{c|c|c|c|c|c}  
		\hline
		\hline
		Q for testing & 90 & 70 & 50 & 30 & 10 \\
		\hline
		OEDS  & 36.08\% & 38.83\%   & 38.11\% & 44.58\% & 49.01\% \\
		TSR   & 14.20\% & 27.21\%   & 33.72\% & 39.88\% & 46.59\% \\
		TSR-S & 5.51\%  & 11.88\%    & 15.94\% & 31.32\% & 42.28\% \\
		MBRS  & \textbf{0.00}\%  & \textbf{0.00063\%} & \textbf{0.081}\% & \textbf{5.37}\%  & \textbf{28.54\%} \\
		\hline
		\hline
	\end{tabular}
\end{table}

\subsubsection{Comparison with OEDS and TSR}

To show the advantage of the MBRS method, we train our model under OEDS and TSR method for comparison. For OEDS method, we train the model end-to-end with simulated JPEG-Mask distortion, and stop training before the decoding loss against real JPEG compression increases (to prevent it from overfitting JPEG-Mask). For TSR, we follow the two-stage scheme to train the model end-to-end without noise layer at stage one, and train the decoder under real JPEG compression at stage two. As an intuitive combination of OEDS and TSR mentioned in Section 1, we propose the TSR with Simulated JPEG (TSR-S), to replace the noise-free layer with JPEG-Mask in TSR at stage one. For all real JPEG compression, $Q=50$ is applied in training, and different $Q$ is applied in testing. As shown in Table 6, OEDS method leads to weak robustness against real JPEG because of “ovrfitting” of simulated distortion and TSR method shows little higher robustness. The TSR-S scheme reduce the BER by about $17\%$ at $Q=50$, while our MBRS method greatly enhance the robustness so that BER drops to almost $0\%$. This shows the superior of our MBRS method to greatly enhance the robustness against JPEG compression.

\subsubsection{Diffusion Analysis}

In Section 4.2.2, we find that our model is vulnerable to crop and cropout attack, and this is due to the structure of our message processor and decoder. As is shown in Figure 5, the message processor reshape the message to a 2-dimensional tensor and feed it into several convolution layers, so only a small region of the encoded image obtain the information of each bit because of the limitation of the field-of-view. In other word, each bit in the message can only be embedded in pixels that are nearby the position of this bit in the 2-dimension message tensor. Similarly, in the last layer of the decoder we directly reshape the tensor to the decoded message, for which the decoder can only decode each message bit from a small region of image instead of from the whole image. As a result, our model will lose all information of one bit, if all pixels that contain information of this bit are removed or replaced, for example, under crop or cropout distortion. To solve this problem, we add a diffusion and an inverse block in the model. In the message processor, we can use a full connection layer to subject the raw message to another tensor, and then reshape this tensor and feed it into the convolutional layers. Accordingly, we can use a full connection layer to subject the output tensor into the decoded message. With the additive diffusion block, we believe the information of each bit in the message an be diffused to the whole image.\\

The added diffusion blocks are jointly trained with the raw encoder-decoder. To justify our design, we can make a comparison between our model with the diffusion block and \cite{Zhu_2018_ECCV,inproceedings} again. MBRS method utilizes simulated JPEG-Mask, real JPEG ($Q=50$), Crop ($p=0.0225$) and Identity layer. Results are shown in Table 2. With the diffusion block, our model achieves a much lower BER under crop, cropout and dropout distortions, which indicates the diffusion function of the additive blocks. At the same time, our model still maintains strong robustness against JPEG compression and Gaussian filter, which means that our model over-performs the baseline under all of the tested distortions. Besides, to show the effect of the diffusion blocks, we randomly select an image, and embed an all-"0" message into it to obtain $I_{EN_0}$, then change one bit to "1" and embed again to obtain $I_{EN_1}$. After that, we can show the embedding position of this bit by show the residual $diff=|I_{EN_0}-I_{EN_1}|$. The results are shown in Figure 6. We find that with the diffusion block, our model succeed in embedding each bit into the whole image, leading to stronger robustness toward crop attack. 

\begin{figure}[t]
	\includegraphics[width=\linewidth]{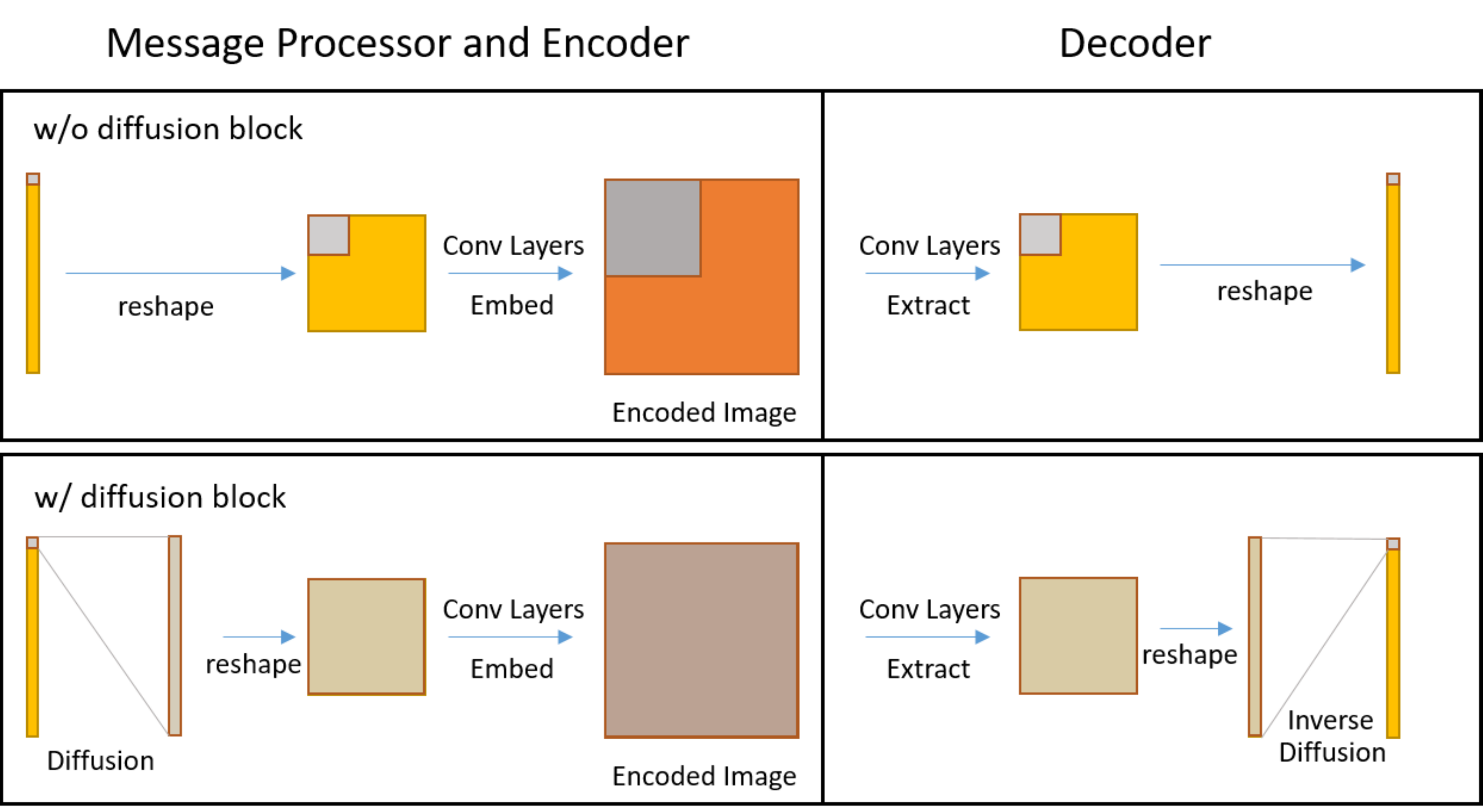}
	\caption{Each bit in the message is embedded and extracted only in a small region of the image that is nearby the position of this bit. We add an additive diffusion block into the model, which utilize a full connection layer to diffuse each bit into the whole image.}
	\Description{The diffusion block.}
\end{figure}

\begin{figure}[t]
	\includegraphics[width=\linewidth]{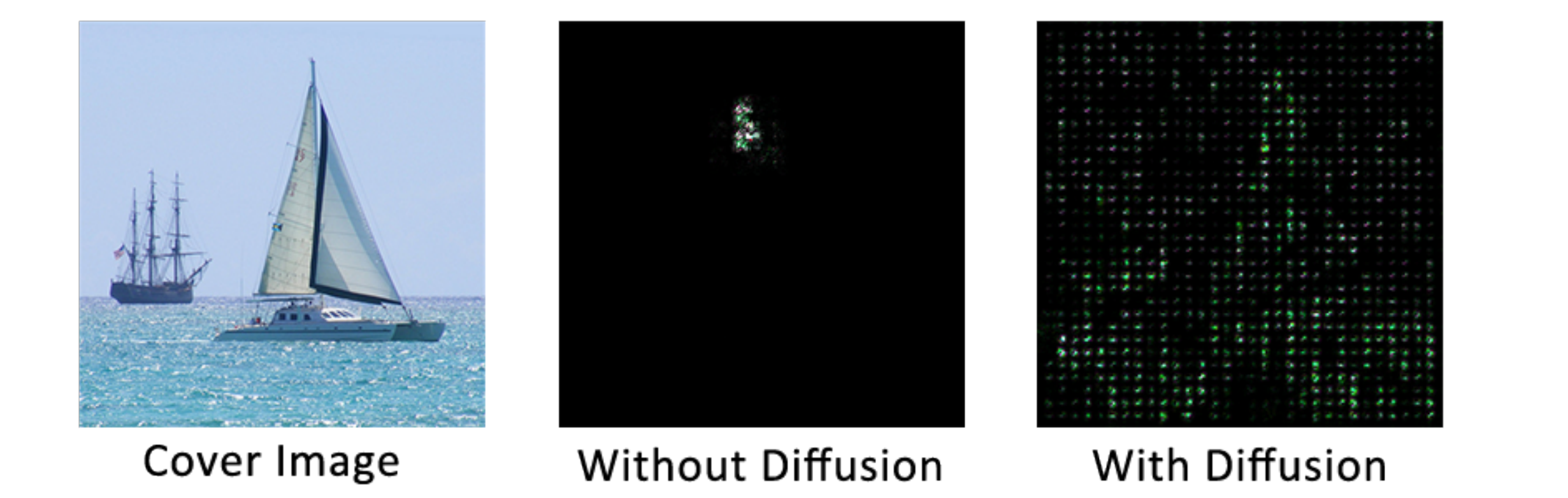}
	\caption{The left image is the cover image to be encoded. We embed an all-"0" message into it, and then change one bit to "1" then embed into it again. Subtract the two image and normalize the diff to [0, 255]. In the middle is the result of the original model. The right image shows the result of the model with diffusion block.}
	\Description{The diffusion block.}
\end{figure}

\section{CONCLUSION}
In this paper, we propose a novel Mini-Batch of Simulated and Real JPEG compression (MBRS) method to enhance robustness against JPEG compression, in which we randomly select one from simulated JPEG, real JPEG and noise-free layer as the noise layer for each mini-batch. To make the best use of our MBRS method, we utilize the Squeeze-and-Excitation block and a message processor for better performance, and an optional diffusion block for robustness against crop attack.  The extensive experiment shows that our method performs better in not only JPEG robustness but also image quality.

\begin{acks}
This work was supported in part by the Natural Science Foundation of China
under Grant 62072421, and 62002334, and by Anhui Science Foundation of China under Grant 2008085QF296, and by Exploration Fund Project of University of Science and Technology of China under Grant YD3480002001.
\end{acks}

\bibliographystyle{ACM-Reference-Format}
\bibliography{mbrs}

\end{document}